\documentclass[10pt,twocolumn,letterpaper]{article}

\usepackage{wacv}
\usepackage{times}
\usepackage{epsfig}
\usepackage{graphicx}
\usepackage{amsmath}
\usepackage{amssymb}
\usepackage{enumerate}
\usepackage{enumitem}
\usepackage{algorithm}
\usepackage{algorithmicx}
\usepackage{algpseudocode}
\usepackage{csquotes}
\usepackage{multirow}
\usepackage[normalem]{ulem}
\usepackage{url}

\wacvfinalcopy % *** Uncomment this line for the final submission

\ifwacvfinal
\fi

\ifwacvfinal
\usepackage[breaklinks=true,bookmarks=false]{hyperref}
\else
\usepackage[pagebackref=true,breaklinks=true,colorlinks,bookmarks=false]{hyperref}
\fi

\begin{document}

%%%%%%%%% TITLE
\title{Learning Gaussian Representation for Eye Fixation Prediction}
%A Simple and Fast Eye Fixation Prediction Model by Learning Gaussian Features

\author{Peipei Song\\
Data61/CSIRO, Australian National University\\
Canberra, ACT 2601, Australia\\
\and
Jing Zhang\\
Australian National University\\
Canberra, ACT 2601, Australia\\
\and
Piotr Koniusz\\
Data61/CSIRO, Australian National University\\
Canberra, ACT 2601, Australia\\
\and
Nick Barnes\\
Australian National University\\
Canberra, ACT 2601, Australia\\
}

\maketitle
%\thispagestyle{empty}

%%%%%%%%% ABSTRACT
\begin{abstract}
Existing eye fixation prediction methods perform the mapping from  
input images to the corresponding dense fixation maps generated from raw fixation points. However, due to the stochastic nature 
of human fixation, the generated dense fixation maps
may be a less-than-ideal representation of human fixation. To provide a robust fixation model, we introduce \enquote{Gaussian Representation} for eye fixation modeling. Specifically,
we propose to model the eye fixation map as a mixture of probability distributions, namely a Gaussian Mixture Model. In this new representation, we use several Gaussian distribution components as an alternative to the provided fixation map, which makes the model more robust to the randomness of fixation. Meanwhile, we design our framework upon some lightweight backbones to achieve real-time fixation prediction.
Experimental results on three public fixation prediction datasets (SALICON, MIT1003, TORONTO) demonstrate that our method is fast and effective. 
\end{abstract}

\section{Introduction}
\begin{figure}
    \centering
    \includegraphics[width=1.0\linewidth]{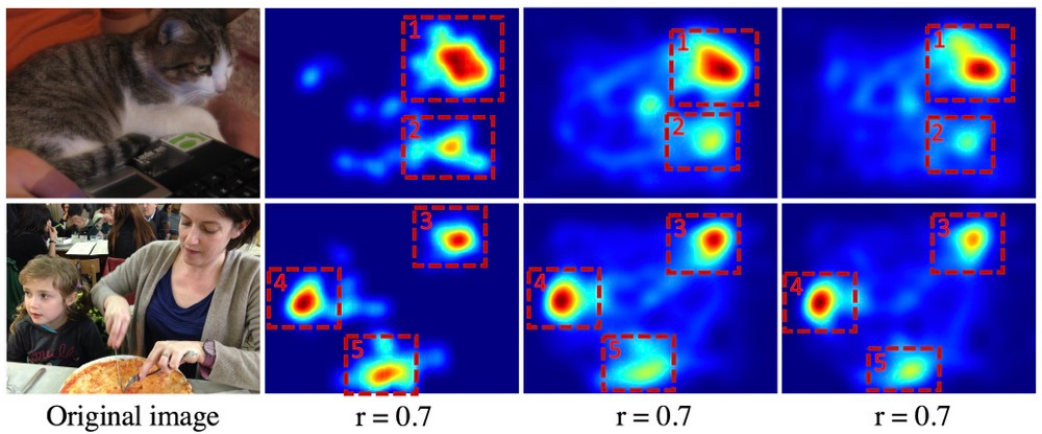}
    \caption{Dense fixation maps with different times of random selection.
    Param. \enquote{r=0.7} indicates that we randomly sample 70\% participants to generate the Gaussian blurred eye fixation map. }
    \label{fig:subjectivity_maps}
\end{figure}

While free viewing a specific scene, our eyes tend to focus on salient and informative regions that attract our attention. As a low-level image processing technique, saliency detection plays an important role in human visual recognition. 
Computer vision focuses on two main problems for saliency: eye fixation prediction
and salient object detection. 
The former produces a heat map representing the degree of attention while the latter generates a salient segmentation map that highlights the entire scope of the salient object.
As shown in Fig.~\ref{fig:subjectivity_maps}, in contrast to salient object detection, 
fixation prediction is usually defined as a regression task that localizes the most discriminative region that attracts human attention.

The human visual system enjoys 
two kinds of attentional mechanisms: bottom-up \cite{firsteyefixation, itti1998model, harel2007graph, covsal, AIM, zhang2013saliency, cas} and top-down \cite{vig2014large, deepgaze1, deepgaze2, pan2017salgan, pan2016shallow, kruthiventi2017deepfix, cornia2018predicting, dodge2018visual, cornia2016deep, liu2018deep, jetley2016end, jia2020eml, fosco2020much}. 
Bottom-up methods, also called feature driven methods, search for conspicuous regions in the image by computing contrast features, such as brightness, color, orientation, optical flow,  \etc. 
Top-down data-driven methods learn saliency from a large number of training examples \cite{Salicon} with clean labels. Such methods usually involve a deep convolutional neural network, which achieve fixation prediction by regression to a dense fixation map.

\begin{figure}
\label{fig:speed_size}
    \centering
    \includegraphics[width=0.98\linewidth]{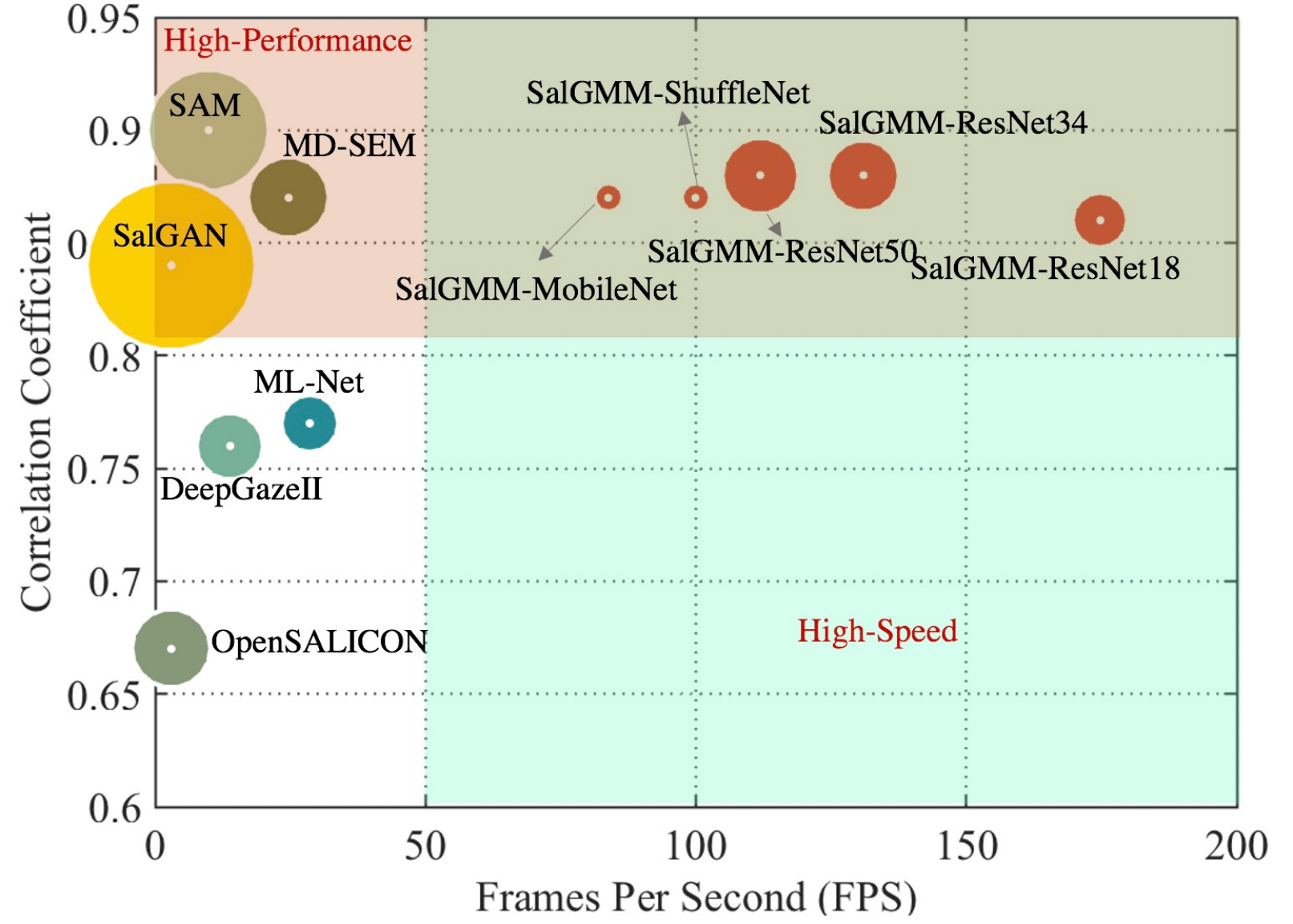}
    % \vspace{-1mm}
    \caption{Comparison of state-of-the-art methods w.r.t. the number of model parameters, inference speed and  
    the performance (correlation coefficient). The size of the circles indicates the number of model parameters. \enquote{SalGMM-*} are several of our models where \enquote{*} should be replaced by specific names of backbone networks.
    %\NB{Could do High-speed from 50FPS is probably reasonable. What do the arrows mean?}
    }
\end{figure}

In the standard pipeline of the top-down models, the raw fixation points are blurred with a Gaussian filter to produce a dense fixation map serving as ground truth for fixation prediction, where the fixation points are generally obtained by recording the fixation locations of multiple participants as they view the image.
We observe that there exists a significant diversity in eye fixation patterns across individuals, as shown in Fig.~\ref{fig:subjectivity_maps}.
For images from the SALICON \cite{Salicon} dataset, we randomly select 70\% of participants and then obtain the dense eye fixation map on
the selected eye fixation points. With three random selections, we obtain three different fixation maps in Fig. \ref{fig:subjectivity_maps}.

The first example in Fig.~\ref{fig:subjectivity_maps} has two main focus points (the cat head and the laptop keyboard) as shown in the marked bounding boxes 1 and 2. The second example contains three regions of focus (the boy's head, the woman's head, and the pizza) as indicated by the marked bounding boxes 3, 4, and 5. While the three groups of participants agree on the main salient regions, the shape of the activation areas varies between individual participants. Meanwhile, there exist some isolated outliers (points outside of the boxes, particularly in columns 2 versus 3 and 4).
This poses a question: 
\enquote{is reconstructing a per-pixel accurate fixation map the only way to model fixation prediction?}

As a regression task, the pixel-wise regression loss
is usually used as a
similarity measure for predicting the eye fixation maps, we argue that the randomness of the dense fixation map (as shown in Fig. \ref{fig:subjectivity_maps}) affects the generalization ability of the regression task, as fitting the isolated outliers may lead to unstable training. Therefore, we propose to model the dense fixation map with the Gaussian Mixture Model. In this way, instead of predicting a pixel-wise map, we operate in the parametric space of a distribution captured by the Gaussian Mixture Model (GMM) which represents a fixation map. We know that
the process of annotation is subjective 
and thus the generated fixation maps are suboptimal from the statistical point of view due to variations in the behavior of different participants. In contrast, predicting parameters of a probability distribution requires fewer variables to predict which reduces overfitting. Thus, we learn a set of probability distributions rather than 
the fixation maps constructed from ground-truth fixation points.

Moreover, eye fixation prediction is usually defined as a pre-processing step for other computer vision-related fields,
including video compression~\cite{hadizadeh2013saliency}, object and action recognition \cite{walther2002attentional,rapantzikos2009dense}, \etc.
Therefore, the processing speed becomes a bottleneck for practical application use. In contrast, learning from a set of GMMs rather than dense fixation maps  requires a relatively smaller network as shown in Fig.~\ref{fig:speed_size}. Our model with ResNet18 \cite{ResNet} as the backbone (\enquote{SalGMM-ResNet18}) achieves comparable performance to the state-of-the-art model (MD-SEM \cite{fosco2020much}) while being seven times faster. We also show that employing lightweight backbones such as ShuffleNet \cite{ma2018shufflenet} in our approach (\enquote{SalGMM-ShuffleNet}) leads to a network that is fast and small, and the obtained results are close to the state of the art. This potentially makes high-quality saliency prediction realistic on mobile devices \cite{ma2018shufflenet} and edge devices \cite{sonyAI}.

%%%% contribution
To account for the randomness of labeling and provide an efficient eye fixation prediction model, we design an end-to-end  neural network learning Gaussian Mixture Models, where dense fixation maps are represented as GMMs. To this end, a reconstruction loss is designed to learn the parameters of GMMs obtained from ground-truth eye fixation points.
Our contributions are summarized as follows:
\renewcommand{\labelenumi}{\roman{enumi}.}
\vspace{-0.2cm}
\hspace{-1.0cm}
\begin{enumerate}[leftmargin=0.6cm]
    \item We formulate 
    the eye fixation map as a Gaussian Mixture Model. In contrast to existing methods, we employ a deep network to predict the parameters of a GMM instead of a dense saliency map per input image. For spatial locations, we express them w.r.t. the nearest reference centers represented by a spatial grid imposed over the spatial domain of images.
    \vspace{-0.2cm}
    \item With the use of annotations for
    eye fixation, we design a novel reconstruction loss for learning the parameters of a GMM. 
    \vspace{-0.2cm}
    \item We perform ablation studies to investigate the impact of various backbones and spatial grid layouts and obtain real-time performing models as shown in Figure \ref{fig:speed_size}.
    \vspace{-0.2cm}
    \item Experimental results on three benchmark fixation prediction datasets show both the effectiveness and efficiency of our solution.
\end{enumerate}

\section{Related Work}
Below, we briefly introduce existing eye
fixation prediction models and GMM-related eye fixation models.

\vspace{0.05cm}
\noindent\textbf{Bottom-Up Models.}
Much research has been published on eye fixation prediction since the first paper \cite{firsteyefixation} in 1985. Approach \cite{tsotsos1995modeling} proposed to model visual attention via selective tuning based on a hypothesis for primate visual attention. Following \cite{tsotsos1995modeling}, Itti \etal~\cite{itti1998model} presented  a visual attention system that combines multi-scale image features into a single topographical saliency map. This feature based model is robust to complex natural scenes. However, it cannot learn from new data because the model used hand-crafted features including color, intensity, and orientation.  As in approach \cite{itti1998model}, GBVS is a model consisting of two steps: forming an activation map on one feature map and normalizing them before combination \cite{harel2007graph}. CovSal \cite{covsal} paid attention to integration by the use of correlation of image features as meta-features for saliency estimation. In \cite{zhang2013saliency}, a novel  saliency model based on a Boolean map was proposed. They used the concept of boolean map theory to characterize an image by a set of binary images. AIM \cite{AIM} presented a novel visual saliency model
on Information Maximization 
to explain the behaviors and reasons for certain components in visual saliency.

The aforementioned models are strong performers on various complex natural scenes because they have a simple architecture and are feature-based. In recent years, deep learning has become popular in computer vision. Deep models for eye fixation prediction  significantly improve the performance compared with the traditional methods. Thus, we detail them below.

\noindent\textbf{Top-Down Models.}
An early attempt at predicting saliency with a convnet was the ensemble of Deep Networks (eDN) \cite{vig2014large}.
This approach inspired DeepGaze I \cite{deepgaze1} to adopt a deeper network. In particular, DeepGaze I used the existing AlexNet network \cite{krizhevsky2012imagenet}, where the fully connected layers were removed to keep the feature maps from the convolutional layers. The responses of each layer were fed into a linear model to learn weights. DeepGaze I is the first case of fine-tuning a convnet trained for classification whilst used for saliency. 
SalGAN adopts a second network discriminator to discriminate the output of a regular eye fixation generator with ground truth \cite{pan2017salgan}. Approach \cite{pan2016shallow} proposed two designs of network: a shallow convnet and a deep convnet, and formulated the eye fixation prediction as a minimization of the Euclidean distance of the predicted saliency map with the provided ground truth. This paper demonstrates that adversarial training  improves the performance of saliency models. Approach \cite{liu2018deep} is a novel deep spatial contextual LSTM model.
SAM \cite{cornia2018predicting} is also an LSTM based saliency attentive model that iteratively refines the predicted saliency map. Following SAM \cite{cornia2018predicting}, MD-SEM \cite{fosco2020much} also adopted LSTM in their network to model the multiple duration saliency.  
 ML-Net \cite{cornia2016deep} combined features extracted at different levels of a Convolutional Neural Network (CNN) for eye fixation prediction.
Approach \cite{wang2018salient} proposed an attentive salient network that  learns the segmentation of salient objects from fixation maps under the assumption that  salient object segmentation maps are fine-grained object-level representations of fixation maps.
Approach \cite{wang2017deep} introduced a multi-scale eye fixation prediction model based on VGG-16 \cite{VGG}, with side outputs from lower levels fused together to achieve multi-scale eye fixation prediction. 
Approach \cite{mrcnn} uses a multi-resolution CNN for predicting eye fixation. Each of these CNNs is trained to classify image patches as fixation-centered or non-fixation-centered.
 Approach \cite{kruthiventi2017deepfix} proposes a one-stream eye fixation framework with a location-biased convolutional layer to preserve center bias during the network training and two inception modules to obtain multi-scale representation.

 Eye fixation prediction methods \cite{jetley2016end, cornia2018predicting} that formulate the eye fixation map as the probability distribution use a dedicated  distribution-related loss. Approach \cite{jetley2016end}
formulated an eye fixation map as a generalized Bernoulli distribution and then trained a deep model to predict such maps using novel loss functions. SAM \cite{cornia2018predicting} calculated the Correlation Coefficient and KL divergence as the loss functions.

%%%%%%%%%%%%%%%%%%%
\vspace{0.05cm}
\noindent\textbf{GMM-related Models.} A Gaussian Mixture Model is a probabilistic model based on a mixture of Gaussian distributions from observed data. 
For the classification problem, approach \cite{variani2015gaussian} directly estimated a GMM with a neural network to learn discriminative features by jointly optimizing the DNN feature layer and the GMM layer.  Approach \cite{zhu2020cross} proposed a patch-based model with shared latent variables from a GMM. 
For the eye fixation prediction, approach \cite{cornia2018predicting} only learns a set of center-bias prior maps generated with Gaussian functions. Several papers
\cite{xu2015learning, xu2017saliency, ren2017learning} used bottom-up hand-crafted features using the model \cite{itti1998model}, and modeled the corresponding eye fixation data using a GMM. In contrast to our approach, none of these models do not incorporate end-to-end learning of GMM features. 

Approach \cite{yonetani2012multi} employed GMM to fit the gaze points from the predicted saliency map for scan path estimation. Method \cite{guo2011improved} estimated a GMM from the feature maps extracted from a bottom-up model referred to as Itti \cite{itti1998model}.  In contrast, our method learns to predict fixation maps in the parametric space of GMM from a given image by using CNN.

Our proposed approach is the first to use an end-to-end neural network to learn the parameters of a GMM to predict a probability distribution function over the image, that is the saliency map which predicts fixation.

\section{Method}
In this section, we first show that a fixation map can be represented by a 2D GMM with little error. We then explain how to use the neural network to learn a GMM with the supervision of  
eye fixation annotations.

\setlength{\tabcolsep}{1.4pt}
\begin{table}
\footnotesize
\begin{center}
\centering
\begin{tabular}{l|ccccc}
\hline
Dataset & Amount & Annotation & Viewers & Points & Resolution\\
\hline \hline
SALICON\cite{Salicon}&20,000&mouse clicking&49-87&460&$480\times 640$\\

MIT1003\cite{judd2009learning}&1003&eye tracking&15&66&Not fixed  \\ 

TORONTO\cite{bruce2007attention}&120&eye tracking&20&93&$511\times 681$ \\
\hline
\end{tabular}
\vspace{3mm}
\caption{Summary of the eye fixation datasets.}
\vspace{-4mm}
\label{tab:dataset_details}
\end{center}
\end{table}

\subsection{Fixation dataset analysis}
\label{gmm_possibility}
\begin{figure}
    \centering
    \includegraphics[width=1.0\linewidth]{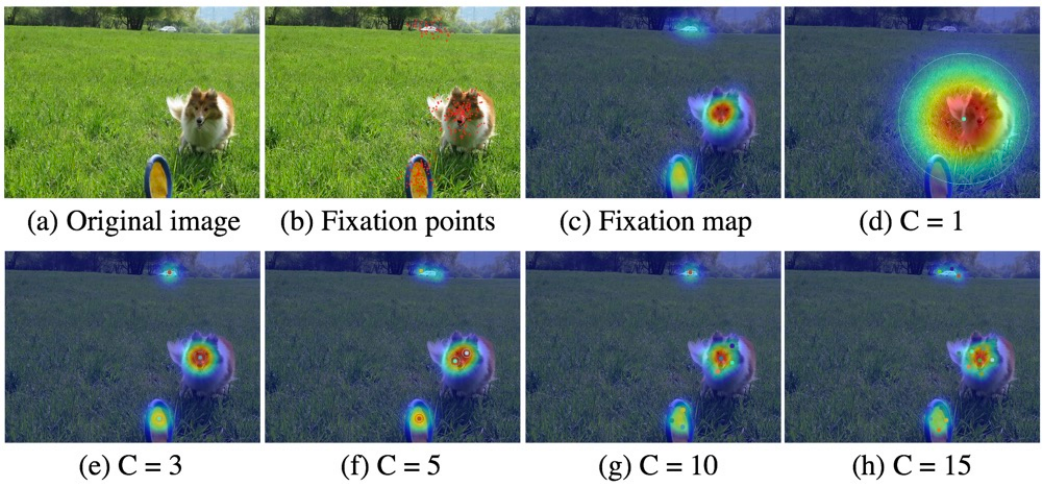}
    \caption{
    Visualization of the image, annotations, and GMM fitted fixation maps. 
    Fixation map (c) is the Gaussian blurred eye fixation map. Figures (d)-(h) show the reconstructed fixation map given a different number of Gaussian components $C$.
    The solid circle and radius of the outer circle are the mean and standard deviation of the fitted Gaussian distribution.
    }
    \label{fig:vis_gmm}
\end{figure}

While most eye fixation prediction datasets use eye tracking \cite{judd2009learning,bruce2007attention} as shown in Table \ref{tab:dataset_details}, the SALICON dataset \cite{Salicon} validated mouse clicks as a model for eye fixation. The protocol based on mouse clicks  enabled SALICON to be the largest eye fixation dataset.
The original fixation ground truth contains
the eye fixation points, as shown in Fig.~\ref{fig:vis_gmm}(b). 
The dense fixation ground truth
is composed by taking the aggregated fixation points from multiple viewers and
applying a 2D Gaussian blur operation on the fixation points, as shown in Fig.~\ref{fig:vis_gmm}(c). By analyzing multiple datasets, we find that fixation points fall around key image locations. For example, in Fig.~\ref{fig:vis_gmm}(b), the fixation points are located on the dog, frisbee, and car. Therefore, the saliency map can be regarded as a Probability Density Function (PDF) of the saliency over spatial locations in the image, and it can be captured with the Gaussian Mixture Model.

\vspace{2mm}
\noindent \textbf{Fitting Eye Fixation Map via GMMs.}
\vspace{1mm}

Firstly, we will analyze the possibility of using a GMM \cite{reynolds2009gaussian} to parameterize the eye fixation map. Let us take the SALICON dataset as an example.
The fixation points are described as $\boldsymbol{P}\equiv\{\boldsymbol{p}^k\}_{k=1}^N$,
where $N$ is number of fixation points, $k$ is an index over fixation points, and $\boldsymbol{p}^k = (u^k,v^k)$
is the coordinate of the fixation point. In the standard fixation prediction pipeline \cite{Salicon}, the ground truth fixation map $\boldsymbol{I_{gt}}$ is obtained by applying a 2D Gaussian blur (for SALICON, $\sigma=19$) on $\boldsymbol{P}$. Instead of directly regressing w.r.t. $\boldsymbol{I_{gt}}$, we first obtain GMM on the fixation points with a parameter set
$\boldsymbol{\Theta}\equiv\{(\pi^c, \mu_u^c, \mu_v^c, \sigma^c_u, \sigma_v^c, \sigma^c_{u,v})\}^{C}_{c=1}$, where
$\pi^{c}$ is the weight for each Gaussian component, $\boldsymbol{\mu^{c}}= (\mu^c_u, \mu^c_v)$ is the location of Gaussian component, $\boldsymbol{\Sigma^{c}}$ is the covariance matrix and $c$ indexes the Gaussian components. The Gaussian covariance matrix is simply given as:
\begin{align}
    \boldsymbol{\Sigma^c}= 
    \left [
    \begin{matrix}
    \sigma_u^c & \sigma_{u,v}^c  \\
    \sigma_{u,v}^c & \sigma_v^c 
    \end{matrix}
    \right ].
    \label{eq:Gaussian_cov}
\end{align}

We then reconstruct one fixation map $\boldsymbol{I^c}$ from the estimated component as:
\begin{align}
    \boldsymbol{I^c}\equiv\boldsymbol{I}(\boldsymbol{p}; \boldsymbol{\mu^c},\! \boldsymbol{\Sigma^c})=\frac{1}{2\pi \sqrt{|\boldsymbol{\Sigma^c}|}}e^{-\frac{1}{2}(\boldsymbol{p}-\boldsymbol{\mu^c})^T {\boldsymbol{\Sigma^c}}^{^{-1}} (\boldsymbol{p}-\boldsymbol{\mu^c})},
    \label{eq:norm_dist}
\end{align}
\noindent{where
$\boldsymbol{p}$} is the coordinate of fixation point. 
With multiple Gaussian components, we obtain the reconstructed fixation map $\boldsymbol{I} = \sum_{c=1}^C \pi^c \boldsymbol{I}^c$ (we drop  parameters for brevity).

Below, we set a different number of Gaussian components to fit the fixation points. Fig.~\ref{fig:vis_gmm}(d)-(h) illustrates the reconstructed fixation map obtained by $C$ Gaussian components. Fig.~\ref{fig:vis_gmm} shows that the reconstructed fixation map with $C=3$ has three separate focus locations on dog, frisbee, and car. When using more Gaussians, \eg $C=5$, to fit the fixation points, some Gaussian components may overlap with each other to fit the shape of fixation points. This observation motivates us to use relatively larger numbers of Gaussian components to obtain a better representation.

Finally, we compare the reconstructed eye fixation map $\boldsymbol{I}$ with the ground truth dense fixation map $\boldsymbol{I_{gt}}$. 
For the number of Gaussian components set to 20, we obtain  Mean Squared Error (MSE), Kullback-Leibler divergence (KL), Correlation Coefficient (CC), Similarity (SIM), and Normalized Scanpath Saliency (NSS) equal to 0.0016, 0.0783, 0.9809, 0.8704, 2.3530 respectively. 
This shows that a GMM can be used to parameterize the eye fixation map with little error.

Thus, a saliency map can be modeled as a GMM over the image space. Rather than using a neural network to model the PDF directly by learning an image map using regression, in this paper, we directly output parameters of a GMM from our neural network.

\subsection{Our proposed network-SalGMM}

\begin{figure*}
    \centering
    \includegraphics[width=0.90\linewidth]{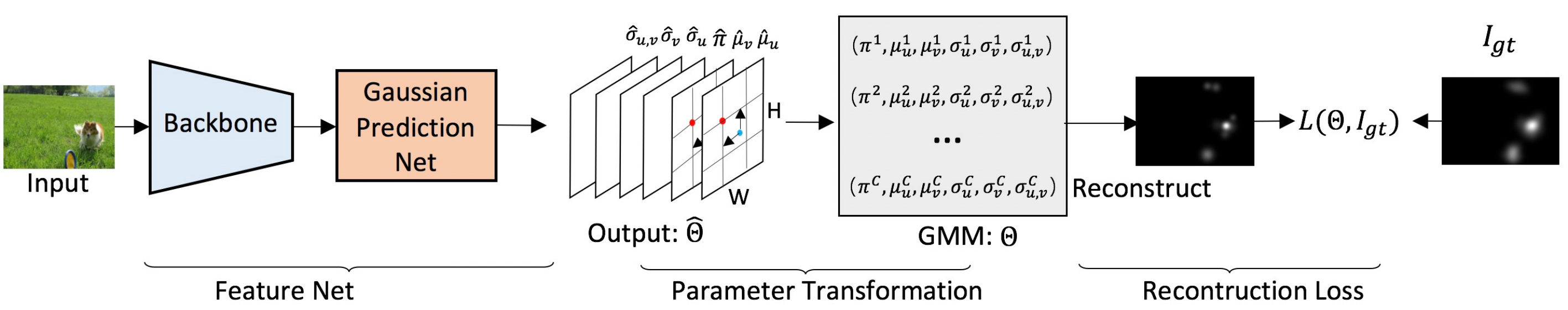}
    \caption{Network architecture. Note that centers of Gaussians are expressed w.r.t. the spatial reference grid.}
    \label{fig:network}
\end{figure*}

Section \ref{gmm_possibility} demonstrated that it is possible to use a GMM in place of a dense eye fixation map.
In this section, we build a neural network to learn a GMM for eye fixation prediction, as shown in Fig.~\ref{fig:network}. Our network includes three main parts: the \enquote{Feature Net}, \enquote{Parameter Transformation}, and \enquote{Reconstruction Loss}.  
 
\subsubsection{Feature Net}
As models that are fine-tuned on the ImageNet \cite{krizhevsky2012imagenet} perform better than those that are trained from scratch, we adopt commonly used feature extraction networks as the backbone. Following the backbone network, the \enquote{Gaussian Prediction Net} in Fig. \ref{fig:network} estimates the parameters of GMM. We only use \enquote{Pooling} and \enquote{Convolutional} layers in the Gaussian Prediction Net to preserve spatial information.
In the Gaussian Prediction Net, we combine multi-scale features by using a similar architecture to FPN \cite{lin2017feature}. The outputs of Feature Net is a $H\times W\times K$ dimensional feature map, where each $K$ dimensional vector inside the feature map represents one Gaussian component ($K=6$), as shown in Fig.~\ref{fig:network}. The number of components $C=H\times W$ which coincides with the spatial size of feature maps.
In this paper, we investigate
3 down-sampling scales: 32, 64 and 128. The outputs of the 3 scales are 16, 36, and 64 Gaussian components, respectively. 
We will show in the experimental results section that the number of Gaussian components is sufficient for our problem at hand. The output $\boldsymbol{\hat{\Theta}} \in \mathbb{R}^{H\times W\times K}$ represents a Gaussian component, and we define one Gaussian component as 
$\boldsymbol{\hat{\theta}^c}\equiv(\hat{\pi}^{c}, \hat{\mu}_u^{c}, \hat{\mu}_v^{c}, \hat{\sigma}_u^{c},  \hat{\sigma}_v^{c},  \hat{\sigma}_{u,v}^c)$, where $c \in \{1,\cdots,C\}$.

\subsubsection{Parameter Transformation}
\begin{figure}
    \centering
    \includegraphics[width=0.99\linewidth]{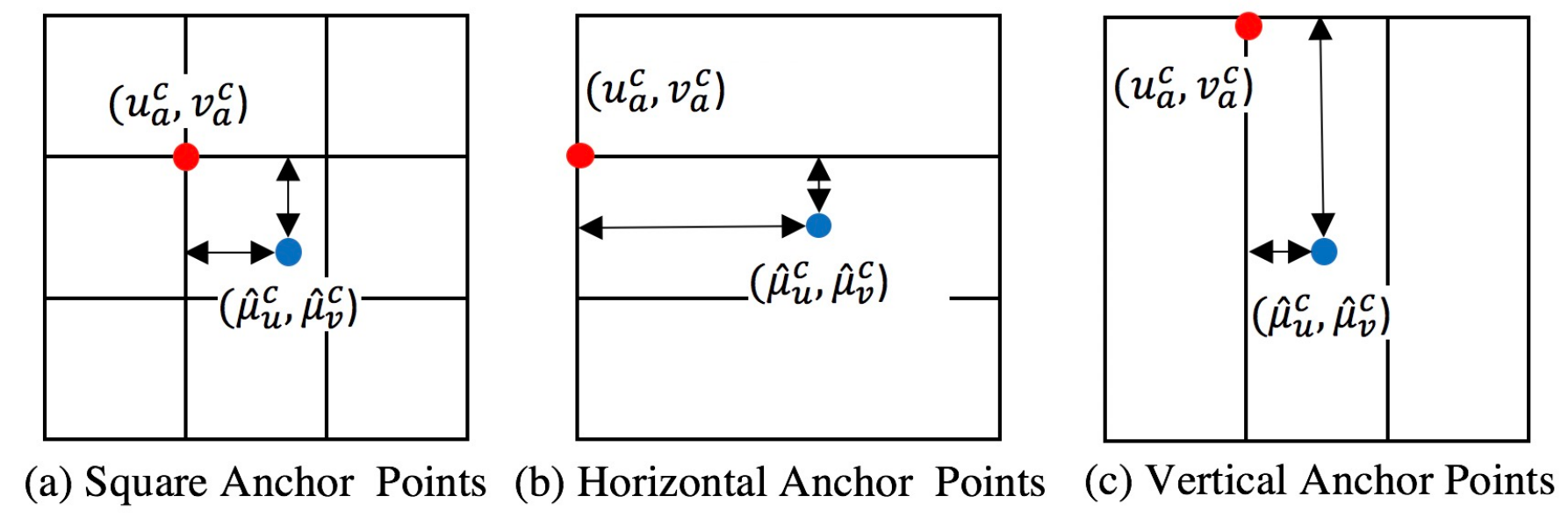}
    \caption{Three kinds of anchor settings (reference points).}
    \label{fig:grid}
    \vspace{-2mm}
\end{figure}

Despite the output of the proposed network is $\boldsymbol{\hat{\Theta}}$, due to the last layer of the \enquote{Gaussian Prediction Net} being a convolutional layer, $\boldsymbol{\hat{\Theta}}$ is not the actual set of GMM parameters that we use to represent the eye fixation map. Inspired by Faster R-CNN \cite{ren2015faster}, we introduce a parameter transformation between the network output $\boldsymbol{\hat{\Theta}}$ and the actual GMM parameters $\boldsymbol{\Theta}$.  
Faster R-CNN
pre-computes multiple spatial anchor boxes w.r.t. the image and uses a neural network to regress the offsets from the anchor box to a nearby ground-truth box. In this paper, to predict a Gaussian component, we regress offsets between the anchor Gaussian centers (reference points) and the ground-truth Gaussian centers. Firstly, we use the \enquote{Sigmoid} activation function to scale the coordinate $(\hat{\mu}^c_u, \hat{\mu}^c_v)$ from the output $\boldsymbol{\hat{\Theta}}$. For example, as illustrated in Fig.~\ref{fig:grid} (a) for { a $3 \times 3$ output map}, the estimated $(\hat{\mu}^c_u, \hat{\mu}^c_v)$ is the offset to the nearby anchor point $(u^c_a, v^c_a)$. The grid size, or the scale of down-sampling in each dimension, is $(w_a, h_a)$. The actual locations of the GMMs are then computed as:

\vspace{-0.3cm}
\begin{align}
    \mu^c_u = (\hat{\mu}^c_u+u^c_a)\cdot w_a \text{ and } \mu^c_v = (\hat{\mu}^c_v+v^c_a)\cdot h_a.
    \label{eq:Gaussian_mean}
\end{align}

In a GMM, all of the $\pi^c$ meet the constraint $\sum_{c=1}^C \pi^c\!=\!1$, thus we use a \enquote{Softmax} layer as the activation function for $\hat{\pi}^c$. The weight $\pi^c = Softmax(\hat{\pi}^c)$, and the covariance $(\sigma^c_u, \sigma^c_v, \sigma^c_{u,v})$ do not require normalization.

\subsubsection{Reconstruction Loss}
\label{loss_function_sec}
\begin{figure}
\centering
\includegraphics[width=1.0\linewidth]{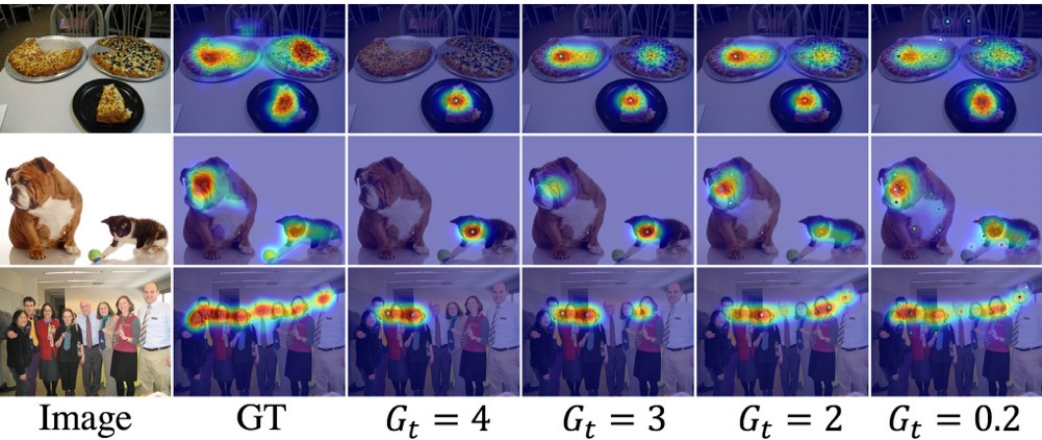}
\caption{Visualization of predictions given different thresholds on the SALICON validation dataset.}
\vspace{-2mm}
\label{fig:vis_salicon_val}
\end{figure}

Given the transformed GMM parameters $\boldsymbol{\Theta}$, we reconstruct the fixation map $\boldsymbol{\hat{I}}$ before we compute the loss function.
We show the pipeline of our method in Algorithm \ref{alg:proposed_algorithm}.  
The reconstructed saliency map is formulated as:
\begin{align}
    \boldsymbol{\hat{I}}=GMM(\boldsymbol{\Theta}\;| \;\pi^c\!>\!G_t/C), c=\{1,\cdots, C\},
    \label{eq:recons}
\end{align}
where
$G_t$ is the threshold for selecting Gaussian components. Fig.~\ref{fig:vis_salicon_val} shows reconstruction maps using different thresholds $G_t$. Our experimental results show that large $G_t$ leads to efficient training and worse performance. The smaller $G_t$, on the contrary,  benefits network performance while leading to longer training time. We set $G_t = 0.2$ in our experiments to achieve a trade-off between network performance and training efficiency.

\begin{algorithm}
    \small
    \caption{Learning GMM Features for Predicting Eye Fixation}
    \label{alg:proposed_algorithm}
    \textbf{Input}: Training images $\boldsymbol{D}=\{\boldsymbol{X_i}, \boldsymbol{I_i}, \boldsymbol{P_i}\}_{i=1}^{N_b}$, where $\boldsymbol{X_i$, $I_i}$ and $\boldsymbol{P_i}$ are input RGB image, its dense fixation map and fixation points map.
    
    \textbf{Output}: Parametric Gaussian Mixture Model $\boldsymbol{\hat{\Theta}}\equiv\{\boldsymbol{\hat{\Theta}_i\}}_{i=1}^{N_b}$. For each input image $\boldsymbol{X}_i$, the output is $\boldsymbol{\hat{\Theta}_i}$, where $C$ is the number of Gaussian components.\\
    \small \textbf{Network Setup: } Maximum epoch is $N_e$, number of batches is $N_b$ and learning rate is $\gamma$. $F$ is the network.
    \begin{algorithmic}[1]
    \State \textbf{Step 1.}  Initialize the backbone with the parameters pretrained on ImageNet.
    \State \textbf{Step 2.} Train the network.
    \For{$t = 1,\cdots,N_e$}
    \For{$i = 1,\cdots,N_b$}
    \State Compute the output of the network: $\boldsymbol{\hat{\Theta}_i} = F(\boldsymbol{X_i})$.
    \State Transform relative coord. to absolute: $\boldsymbol{\hat{\Theta}_i} \rightarrow \boldsymbol{\Theta_i}$.
    \State Reconstruct eye fixation map $\boldsymbol{\hat{I}_i}$ by using $\boldsymbol{\Theta_i}$ (Eq.~\ref{eq:recons}). 
    \State Compute loss according to Eq. \ref{eq:cc_loss}.
    \State Backprop. to update network parameters/learn GMM.
    \EndFor
    \EndFor
    \end{algorithmic}
\end{algorithm}

We define our loss function between the reconstructed saliency map $\boldsymbol{\hat{I}}$ and the eye fixation annotation $\boldsymbol{I_{gt}}$ as:
\begin{align}
    L(\boldsymbol{\Theta}, \boldsymbol{I_{gt}}) =\frac{\sigma(\boldsymbol{I_{gt}}, GMM(\boldsymbol{\Theta}\;| \;\pi^c\!>\! G_t/C))}{\sigma(\boldsymbol{I_{gt}})\cdot\sigma(GMM(\boldsymbol{\Theta}\;| \; \pi^c\!>\!G_t/C))},
    \label{eq:cc_loss}
\end{align}
\noindent
where $\sigma(\cdot)$ evaluates correlation. The correlation coefficient loss is also used in \cite{cornia2018predicting} but in a very different context. As we reconstruct $\boldsymbol{\hat{I}}$ by $\boldsymbol{\Theta}$, the loss will learn parameters of GMM.

\section{Experimental Results}
\subsection{Experimental Setup}

\noindent\textbf{Datasets:}
\textbf{SALICON} \cite{Salicon} contains 10,000 training images, 5,000 validation mages and 5,000 testing images, which are selected from the Microsoft COCO dataset \cite{COCO}.  
The annotations are represented by mouse-clicking instead of eye tracking. We train our model on the SALICON training and validation sets and test on the SALICON test set, \textbf{MIT1003} dataset \cite{judd2009learning} and \textbf{TORONTO} dataset \cite{bruce2007attention}. MIT1003 contains 1,003 images from Flickr and LabelMe, and the TORONTO dataset has 120 color images from indoor and outdoor scenes, as listed in Table.~\ref{tab:dataset_details}.

\vspace{0.05cm}
\noindent\textbf{Training Procedures}: Except where specified, we use ResNet pre-trained on ImageNet as the network backbone. We use two Geforce RTX 2080ti GPUs for training. We freeze the first two ResNet blocks and train the other parameters for 20 epochs with the learning rate $10^{-3}$. The training images are resized to $352\times352$. When testing, the images are also resized to $352\times352$.

\vspace{0.05cm}
\noindent\textbf{Evaluation Metrics:}
The standard evaluation metrics 
are introduced by \cite{bylinskii2018different}, including Pearson’s Correlation Coefficient (\textbf{CC}), Similarity (\textbf{SIM}), Kullback-Leibler divergence (\textbf{KL}), Earth Mover's Distance (\textbf{EMD}), Normalized Scanpath  Saliency (\textbf{NSS}), Area Under Curve (\textbf{AUC}), Information Gain (\textbf{IG}) . 
Three variants for AUC: \textbf{AUC-Judd} \cite{aucbyjudd}, \textbf{AUC-Borji} \cite{borji2012quantitative} and shuffled AUC (\textbf{sAUC}) \cite{borji2012quantitative} are commonly used for evaluation.
Evaluation metrics for eye fixation prediction are divided into two groups according to two kinds of annotations: Distribution-based metrics and Location-based metrics. The blurred fixation map is treated as a probability distribution. The metrics defined on the blurred fixation map are ``Distribution based", including CC, SIM, KL, and EMD. Similarly, the metrics defined on the fixation points are ``Location-based", namely NSS, sAUC, IG, AUC-Borji and AUC-Judd. We compare our proposed model with the competing methods on the two groups of metrics in Section 4.2.

\setlength{\tabcolsep}{2.7pt}
\begin{table}
    \begin{center}
    \scriptsize
    \begin{tabular}{l|c c c|c c c c}\hline
    Methods& \multicolumn{3}{c|}{Distribution-based} & \multicolumn{4}{|c}{Location-based} \\ \cline{2-8}
    ~& CC$\uparrow$ & SIM$\uparrow$ & KL$\downarrow$ & NSS$\uparrow$ & sAUC$\uparrow$ & IG$\uparrow$ & AUC-B$\uparrow$ \\ 
    \hline \hline
    
    Itti\cite{itti1998model}  &0.549&0.543&0.712&1.098&0.629&-0.005&0.773 \\ %run code
    GBVS\cite{harel2007graph}  &0.639&0.585&0.598&1.262&0.612&0.172&0.807 \\ %run code
    %PDP\cite{jetley2016end} &0.765&-&-&-&0.781&-&0.882 \\ %from PDP paper
    OpenSALICON\cite{christopherleethomas2016} &0.657&0.600&0.581&1.553&0.716&0.331&0.807 \\  %SALICON not release code
    %ML-NET\cite{cornia2016deep} &0.743&-&-&2.789&0.768&-&0.866 \\ %from sam paper and salgan, mlnet, DSCLRCN
    ML-NET\cite{cornia2016deep} &0.762&0.668&0.639&1.802&0.723&0.461&0.838 \\ % run code
    SalNet\cite{pan2016shallow} &0.622&-&-&1.859&0.724&-&0.858 \\ %from sam paper and salgan
    %DeepGazeII\cite{kummerer2017understanding}  &0.509&-&-&1.336&0.761&-&0.885 \\ %from sam paper, different numbers are shown in DSCLRCN,sauc=0.787, auc=0.867, nss=1.271, cc=0.479
    DeepGazeII\cite{kummerer2017understanding} &0.762&0.672&0.752&1.742&0.692&0.371&0.845 \\
    DeepFix\cite{kruthiventi2017deepfix} &0.804&-&-&-&0.630&0.315&0.761 \\
    %SalGAN\cite{pan2017salgan} &0.781&-&-&2.459&0.772&-&0.884 \\ %from salgan paper 
    SalGAN\cite{pan2017salgan} &0.842&0.728&\textcolor{red}{0.426}&1.817&0.733&\textcolor{green}{0.650}&0.857 \\
    %DSCLRCN\cite{liu2018deep} &0.831&-&-&3.157&0.776&-&0.884 \\ %from sam paper and salgan paper, DCCLRCN
    %SAM-ResNet\cite{cornia2018predicting}  &0.842&-&-&3.204&0.779&-&0.883 \\ %from sam paper 
    %SAM-ResNet(2015)\cite{cornia2018predicting} &(0.806)&(0.701)&(0.822)&(1.930)&(0.732)&(0.383)&(0.850) \\ %2015 release not give predictions but give code. 
    SAM-Res(17)\cite{cornia2018predicting} &\textcolor{red}{0.899}&\textcolor{red}{0.793}&0.611&\textcolor{green}{1.990}&\textcolor{blue}{0.741}&0.538&\textcolor{green}{0.865} \\ %2017 release not give predictions but give code. AUC: 0.865 CC: 0.899 KL: 0.611 SAUC: 0.741 IG: 0.538 NSS: 1.990 SIM: 0.793. 
    %MixNet\cite{dodge2018visual} &0.730&-&-&2.767&\textcolor{blue}{0.771}&-&0.861 \\  %from sam paper and mixnet paper 
    EML-NET \cite{jia2020eml} &\textcolor{green}{0.886}&\textcolor{green}{0.780}&\textcolor{blue}{0.520}&\textcolor{blue}{2.050}&\textcolor{red}{0.746}&\textcolor{red}{0.736}&\textcolor{red}{0.866} \\
    MD-SEM\cite{fosco2020much} &0.868&0.774&0.568&\textcolor{red}{2.058}&\textcolor{green}{0.745}&0.598&\textcolor{blue}{0.864} \\ %AUC: 0.864, CC: 0.870, KL: 0.623, SAUC: 0.745, IG: 0.598, NSS: 2.059, SIM: 0.775
    \hline
    SalGMM &\textcolor{blue}{0.883}&\textcolor{blue}{0.777}&\textcolor{green}{0.463}&1.898&0.732&\textcolor{blue}{0.603}&0.861 \\
    % Our\_Gap &-0.016&-0.016&-0.037{0.463}&1.898&0.732&\textcolor{blue}{0.603}&0.861 \\
    %CC: 0.883 KL: 0.463 SAUC: 0.732 IG: 0.603 NSS: 1.898 SIM: 0.777
    \hline
    \end{tabular}
    \vspace{2mm}
    \caption{Comparison results on the SALICON Saliency Prediction Challenge (LSUN 2017) testset.
    % The predictions of Itti, GBVS and OpenSALICON are generated by the public code and model. 
    }
    \vspace{-2mm}
    \label{tab:salicon-test-performance}
    \end{center}
\end{table}

\setlength{\tabcolsep}{1.3pt}
\begin{table}
    \begin{center}
    \scriptsize
    \begin{tabular}{l|cccc|ccccc}\hline\noalign{\smallskip} 
    %% sAUC IG EMD use a baseline map
    Methods& \multicolumn{4}{c|}{Distribution-based} & \multicolumn{5}{|c}{Location-based} \\ \cline{2-10}
     & CC$\uparrow$ & SIM$\uparrow$ & KL$\downarrow$ & EMD$\downarrow$ & NSS$\uparrow$ & sAUC$\uparrow$ & IG$\uparrow$ & AUC-J$\uparrow$ & AUC-B$\uparrow$ \\ \hline \hline
    Itti \cite{itti1998model} &0.33&0.32&1.48&5.18&1.10&0.64&0.23&0.78&0.76   \\
    GBVS \cite{harel2007graph} &0.42&0.36&1.30&4.34&1.38&0.63&0.27&0.83&0.82  \\
    AIM \cite{AIM} &0.26&0.27&-&-&0.82&0.68&-&0.79&0.76   \\ %from dva paper
    Judd Model\cite{judd2009learning} &0.30&0.29&-&-&1.02&0.68&-&0.76&0.74   \\ %from dva paper
    CAS \cite{cas} &0.31&0.32&-&-&1.07&0.68&-&0.76&0.74  \\ %from dva paper
    BMS \cite{zhang2013saliency} &0.36&0.33&1.46&5.33&1.25&0.69&0.25&0.79&0.76   \\ %from dva paper, the results are close to our testing with the predictions given. our testing: auc-judd = 0.79, auc-borji=0.77, sauc=0.68, sim=0.33, emd=5.33, cc=0.36, nss=1.25, kl=1.46, ig=0.25\\
    eDN \cite{vig2014large} &0.41&0.30&1.55&5.32&1.29&0.66&0.07&0.85&\textcolor{blue}{0.84}  \\%from dva paper, the results are close to our testing.  prediction are given, our testing: kld: 1.5453, cc: 0.4096, sim: 0.2976, nss: 1.2969, aucj: 0.8579, aucb: 0.8470, sauc: 0.6194, emd: 5.3198, ig: 0.0743
    Mr-CNN \cite{mrcnn} &0.38&0.35&-&-&1.36&\textcolor{blue}{0.73}&-&0.80&0.77  \\%2015 from dva paper
    ML-Net\cite{cornia2016deep} &0.60&0.48&\textcolor{blue}{1.06}&3.50&2.28&0.72&\textcolor{blue}{0.78}&0.86&0.80 \\ %predictions are generated by code, this is our testing
    %DeepFix \cite{kruthiventi2017deepfix} &[0.72]&[0.54]&-&[1.28]&[2.58]&[0.74]&-&[0.90]&[0.85]  \\ % from deepfix paper, finetune on 900 images of mit1003 and test the other 103 images
    %DVA \cite{wang2017deep} &0.64&0.50&-&\textbf{1.29}&2.26&0.71&-&0.87&0.80 \\ %from dva paper 
    DVA \cite{wang2017deep} &\textcolor{blue}{0.64}&0.50&\textcolor{red}{0.95}&3.89&\textcolor{blue}{2.38}&0.72&\textcolor{green}{0.85}&\textcolor{blue}{0.88}&0.81 \\ %predictiona are given
    DeepGazeII\cite{kummerer2017understanding} &\textcolor{red}{0.71}&\textcolor{red}{0.57}&\textcolor{blue}{1.06}&\textcolor{green}{2.48}&\textcolor{red}{2.86}&0.71&0.76&\textcolor{red}{0.90}&0.79\\ % the results are genenrated by code. paper: auc-borji=0.883, sauc=0.777, nss=2.48, ig=0.98 \\
    SalGAN \cite{pan2017salgan} &0.63&\textcolor{blue}{0.51}&\textcolor{green}{1.00}&3.21&2.21&\textcolor{green}{0.74}&0.68&\textcolor{blue}{0.88}&\textcolor{green}{0.85} \\
    %SAM-VGG(2015)\cite{cornia2018predicting} &(0.83)&(0.67)&(0.77)&(1.72)&(3.17)&0.76&1.05&(0.92)&0.84 %predictions are given.
    %SAM-Res(15)\cite{cornia2018predicting} &0.87&0.71&0.74&1.35&3.34&0.77&1.14&0.93&0.84 \\
    SAM-Res(17)\cite{cornia2018predicting} &\textcolor{green}{0.66}&\textcolor{green}{0.53}&1.23&2.94&2.36&\textcolor{red}{0.75}&0.37&\textcolor{green}{0.89}&\textcolor{red}{0.86} \\
    MD-SEM \cite{fosco2020much}&\textcolor{green}{0.66}&\textcolor{green}{0.53}&1.23&\textcolor{red}{1.13}&\textcolor{green}{2.51}&\textcolor{green}{0.74}&\textcolor{red}{0.98}&\textcolor{red}{0.90}&\textcolor{blue}{0.84} \\%cc: 0.6609, sim: 0.5309, kld: 1.2255, emd: 1.1345, nss: 2.5126, sauc: 0.7412, ig: 0.9808, aucj: 0.8976, aucb: 0.8369
    \hline
    %\textbf{Ours} &0.64&0.48&\textbf{0.90}&3.34&2.25&\textbf{0.74}&0.84&0.88&\textbf{0.86} \\ 
    SalGMM&\textcolor{blue}{0.64}&\textcolor{blue}{0.51}&1.09&\textcolor{green}{1.17}&2.26&\textcolor{green}{0.74}&0.52&\textcolor{green}{0.89}&\textcolor{red}{0.86} \\%cc: 0.6433, sim: 0.5126, kld: 1.0886, emd: 1.1743, nss: 2.2609, sauc: 0.7446, ig: 0.5238, aucj: 0.8940, aucb: 0.8607

    %ours&\textcolor{blue}{0.66}&\textcolor{blue}{0.52}&1.05&\textcolor{red}{1.16}&2.32&\textcolor{green}{0.75}&0.56&\textcolor{green}{0.90}&\textcolor{red}{0.86} \\ %cc: 0.6558, sim: 0.5182, kld: 1.0494, emd: 1.1596, nss: 2.3232, sauc: 0.7457, ig: 0.5563, aucj: 0.8960, aucb: 0.8615%. cat2000 for training
    \hline
    \end{tabular}
    \vspace{2mm}
    \caption{\small Comparison results on the MIT1003 dataset.
    % The predictions of Itti, GBVS, ML-Net, SalGAN and DeepGazeII are generated by the public code and model. The predictions of DVA are provided online.
    } %DeepFix is finetuned on 900 images of MIT1003 and tested on the other 103 images. 
    \vspace{-2mm}
    \label{tab:mit1003-performance}
    \end{center}
    
\end{table}

\begin{table}
    \begin{center}
    \scriptsize
    \begin{tabular}{l|cccc|ccccc}\hline\noalign{\smallskip}  
    Methods& \multicolumn{4}{c|}{Distribution-based} & \multicolumn{5}{|c}{Location-based} \\ \cline{2-10}
     & CC$\uparrow$ & SIM$\uparrow$ & KL$\downarrow$ & EMD$\downarrow$ & NSS$\uparrow$ & sAUC$\uparrow$ & IG$\uparrow$ & AUC-J$\uparrow$ & AUC-B$\uparrow$ \\ \hline \hline
    Itti\cite{itti1998model}&0.48&0.48&0.97&2.88&1.30&0.66&0.32&0.80&0.79 \\
    %Itti\cite{itti1998model} &0.48&0.45&(0.97)&(2.88)&1.30&0.65&(0.32)&0.80&0.80  \\ %from dva paper
    GBVS\cite{harel2007graph}&0.57&0.49&\textcolor{blue}{0.85}&2.29&1.52&0.64&0.31&0.83&\textcolor{blue}{0.82} \\
    %GBVS\cite{harel2007graph}&0.57&0.49&(0.85)&(2.29)&1.52&0.64&(0.31)&0.83&0.83 \\  %from dva paper
    AIM\cite{AIM}&0.30&0.36&-&-&0.84&0.67&-&0.76&0.75 \\ % from dva paper
    Judd\cite{judd2009learning} &0.41&0.40&-&-&1.15&0.67&-&0.78&0.77 \\ %from dva paper
    CAS\cite{cas}&0.45&0.44&-&-&1.27&\textcolor{blue}{0.69}&-&0.78&0.78\\ %from dva paper
    CovSal\cite{covsal}&0.56&0.52&1.23&2.09&1.49&0.53&-0.15&0.82&0.77 \\ %predicted maps are given
    BMS\cite{zhang2013saliency} &0.52&0.45&0.98&3.11&1.52&\textcolor{green}{0.71}&0.41&0.80&0.78\\ %predicted maps are given
    %eDN\cite{vig2014large}&0.50&0.40&(1.12)&(3.06)&1.25&0.62&(0.12)&0.85&0.84 \\ %from dva paper
    eDN\cite{vig2014large}&0.50&0.40&1.12&3.06&1.25&0.62&0.12&0.85&\textcolor{red}{0.84} \\ %predicted maps are given
    Mr-CNN\cite{mrcnn}&0.49&0.47&-&-&1.41&\textcolor{green}{0.71}&-&0.80&0.79  \\ %from dva paper
    ML-Net\cite{cornia2016deep} &0.65&0.56&1.12&2.07&2.00&0.69&0.20&0.85&0.76 \\ %generate by code
    %DVA\cite{wang2017deep} &0.72&0.58&(0.65)&(2.18)&2.12&0.76&(0.66)&0.86&0.86 \\  %from dva paper
    DVA\cite{wang2017deep} &0.72&0.58&\textcolor{red}{0.65}&2.18&\textcolor{blue}{2.12}&0.69&\textcolor{blue}{0.66}&\textcolor{blue}{0.86}&0.78  \\ % the predicted maps are given. I test it with MIT benchmark code
    DeepGazeII\cite{kummerer2017understanding}&\textcolor{red}{0.77}&\textcolor{red}{0.65}&1.01&\textcolor{red}{1.48}&\textcolor{red}{2.26}&0.69&0.21&\textcolor{red}{0.88}&0.79\\
    SalGAN\cite{pan2017salgan} &0.72&\textcolor{blue}{0.61}&0.95&1.80&2.02&\textcolor{red}{0.72}&0.27&\textcolor{blue}{0.86}&\textcolor{blue}{0.82} \\
    %SAM-VGG(2015)\cite{cornia2018predicting} &(0.69)&(0.59)&(2.00)&(1.94)&(2.14)&(0.68)&(-1.02)&(0.86)&(0.74) \\
    %SAM-Res(15)\cite{cornia2018predicting} &0.69&0.59&2.03&1.96&\textcolor{blue}{2.12}&0.67&-1.07&\textcolor{blue}{0.86}&0.74 \\
    SAM-Res(17)\cite{cornia2018predicting}&\textcolor{green}{0.74}&\textcolor{green}{0.62}&1.86&1.76&\textcolor{green}{2.14}&\textcolor{red}{0.72}&0.60&\textcolor{blue}{0.86}&\textcolor{blue}{0.82} \\
    MD-SEM \cite{fosco2020much}&0.70&\textcolor{blue}{0.61}&1.82&\textcolor{blue}{1.75}&\textcolor{blue}{2.12}&\textcolor{blue}{0.70}&\textcolor{red}{0.72}&\textcolor{blue}{0.86}&0.78 \\ %cc: 0.6996, sim: 0.6080, kld: 1.8162, emd: 1.7531, nss: 2.1249, sauc: 0.6967, ig: 0.7192, aucj: 0.8616, aucb: 0.7780
    \hline 
    %Res50\_64(30 epochs)&0.73&0.62&1.18&1.70&2.04&0.72&0.63&0.86&0.83 \\  
    SalGMM&\textcolor{blue}{0.73}&\textcolor{blue}{0.61}&\textcolor{green}{0.77}&\textcolor{green}{1.74}&2.06&\textcolor{red}{0.72}&\textcolor{green}{0.69}&\textcolor{green}{0.87}&\textcolor{green}{0.83} \\  % cc: 0.7324, sim: 0.6119, kld: 0.7739, emd: 1.7355, nss: 2.0569, sauc: 0.7177, ig: 0.6902, aucj: 0.8668, aucb: 0.8279. ResNet50\_D64
    \hline
    \end{tabular}
    \vspace{2mm}
    \caption{\small Comparison results on the TORONTO dataset.
    % The predictions of ML-Net, DeepGazeII, SalGAN and SAM are generated by the public code and model. The predictions of CovSal, BMS, eDN and DVA are provided online.
    }
     \vspace{-3mm}
    \label{tab:toronto_performance}
    \end{center}
    
\end{table}

\setlength{\tabcolsep}{5.5pt}
\begin{table*}
    \begin{center}
    \scriptsize
    \begin{tabular}{l|cccc|ccccc|cc}\hline 
    Methods & \multicolumn{4}{c|}{Distribution-based} & \multicolumn{5}{|c|}{Location-based} & \multicolumn{2}{|c}{Model Size} \\ \cline{2-12}
    & CC$\uparrow$ & SIM$\uparrow$ & KL$\downarrow$ & EMD$\downarrow$ &  NSS$\uparrow$ & sAUC$\uparrow$ & IG$\uparrow$ & AUC-J$\uparrow$ & AUC-B$\uparrow$ & Params(M) & FPS\\ \hline \hline
    ML-Net\cite{cornia2016deep} &0.766&0.671&0.567&1.345&1.779&0.674&0.575&0.846&0.740&15.5&28.6\\  %mit matlab
    DeepGazeII\cite{kummerer2017understanding}&0.760&0.673&0.666&1.440&1.703&0.646&0.486&0.854&0.736&20.4&13.9 \\
    %SAM(2015)\cite{cornia2018predicting}&0.810&0.704&0.711&1.268&1.900&0.665&0.528&0.865&0.725&-&-&9.78 \\ %2015 release 
    
    OpenSALICON\cite{christopherleethomas2016}&0.666&0.600&0.562&1.762&1.543&0.692&0.406&0.814&0.758&29.4&3.0 \\ 
    SalGAN\cite{pan2017salgan}&0.841&0.725&0.358&1.199&1.786&0.723&0.782&0.859&0.823&536.5&3.0\\ 
    SAM(2017)\cite{cornia2018predicting}&\textcolor{red}{0.900}&\textcolor{red}{0.794}&0.495&\textcolor{red}{0.864}&\textcolor{green}{1.959}&\textcolor{red}{0.728}&0.694&\textcolor{green}{0.869}&0.829&70.1&9.9 \\ %2017 release
    MD-SEM \cite{fosco2020much}&0.873&\textcolor{green}{0.778}&0.536&\textcolor{green}{0.870}&\textcolor{red}{2.025}&0.702&0.722&\textcolor{red}{0.870}&0.784&30.9&24.7 \\ %cc: 0.8728, sim: 0.7775, kld: 0.5356, emd: 0.8700, nss: 2.0245, sauc: 0.7024, ig: 0.7215, aucj: 0.8700, aucb: 0.7841
    %Ours&0.871&0.740&-&-&1.832&-&-&0.859&0.835&-&-&\\
    
    %\hline
    %\multicolumn{13}{|c|}{Dense Prediction} \\
    %\hline
    %ResNet50\_Dense&0.867&0.662&1.137&3.156&1.804&0.725&0.535&0.858&0.842&26.71&10.53&97.05\\ %cc: 0.8668, sim: 0.6615, kld: 1.1373, emd: 3.1564, nss: 1.8037, sauc: 0.7253, ig: 0.5350, aucj: 0.8576, aucb: 0.8421
    \hline
    \multicolumn{12}{c}{Down sampling Scales} \\
    \hline
    ResNet50\_D32\_S\_AS&\textcolor{green}{0.884}&0.765&\textcolor{red}{0.276}&0.992&\textcolor{blue}{1.880}&0.723&\textcolor{green}{0.862}&\textcolor{blue}{0.866}&0.832&26.7&106.6\\ %cc: 0.8836, sim: 0.7647, kld: 0.2761, emd: 0.9922, nss: 1.8799, sauc: 0.7227, ig: 0.8616, aucj: 0.8656, aucb: 0.8320
    ResNet50\_D64\_S\_AS &0.880&0.769&0.371&\textcolor{blue}{0.950}&1.861&\textcolor{green}{0.725}&\textcolor{blue}{0.858}&0.863&\textcolor{blue}{0.833}&26.7&\textcolor{blue}{111.9}\\ % with recons 58.93. cc: 0.8796, sim: 0.7694, kld: 0.3705, emd: 0.9485, nss: 1.8613, sauc: 0.7253, ig: 0.8584, aucj: 0.8633, aucb: 0.8333
    ResNet50\_D128\_S\_AS&0.837&0.734&0.527&1.048&1.736&0.703&0.709&0.850&0.817&26.7&111.5\\%111.4. cc: 0.8371, sim: 0.7337, kld: 0.5267, emd: 1.0476, nss: 1.7359, sauc: 0.7031, ig: 0.7089, aucj: 0.8503, aucb: 0.8170
    
    %Res50\_64\_PA & 0.880&0.771&-&-&1.864&-&-&0.865&0.834&-&-&-\\
    %Res50\_64\_Multi-rates & 0.880&0.770&-&-&1.856&-&-&0.864&0.836&-&-&-\\ 
    %Res50\_64\_Noise &0.866&0.760&-&-&1.834&-&-&0.862&0.832&-&-&-\\ 
    \hline
    \multicolumn{12}{c}{Gaussian Covariance} \\
    \hline
    ResNet50\_D64\_D\_AS & \textcolor{blue}{0.881}&\textcolor{blue}{0.771}&0.365&0.951&1.862&\textcolor{red}{0.728}&\textcolor{red}{0.864}&0.864&\textcolor{red}{0.837}&26.7&109.2\\ %cc: 0.8807, sim: 0.7711, kld: 0.3649, emd: 0.9514, nss: 1.8617, sauc: 0.7275, ig: 0.8640, aucj: 0.8640, aucb: 0.8369
    ResNet50\_D64\_F\_AS & 0.873&0.725&\textcolor{blue}{0.339}&1.290&1.832&0.719&0.728&0.862&\textcolor{green}{0.836}&26.7&108.5\\  %cc: 0.8730, sim: 0.7246, kld: 0.3389, emd: 1.2901, nss: 1.8320, sauc: 0.7188, ig: 0.7282, aucj: 0.8618, aucb: 0.8360
    
    \hline
    \multicolumn{12}{c}{Anchor Settings} \\
    \hline
    ResNet50\_D64\_S\_AN &0.854&0.746&0.354&1.014&1.793&0.713&0.790&0.857&0.823&26.7&\textcolor{blue}{111.9}\\ %cc: 0.8541, sim: 0.7460, kld: 0.3540, emd: 1.0139, nss: 1.7933, sauc: 0.7132, ig: 0.7898, aucj: 0.8573, aucb: 0.8229
    ResNet50\_D64\_S\_AH &0.841&0.732&0.347&1.069&1.748&0.703&0.743&0.854&0.819&26.7&\textcolor{blue}{111.9}\\ %cc: 0.8411, sim: 0.7318, kld: 0.3468, emd: 1.0692, nss: 1.7484, sauc: 0.7033, ig: 0.7427, aucj: 0.8541, aucb: 0.8189s
    ResNet50\_D64\_S\_AV &0.834&0.718&0.350&1.112&1.719&0.698&0.688&0.850&0.821&26.7&\textcolor{blue}{111.9}\\ %cc: 0.8339, sim: 0.7182, kld: 0.3502, emd: 1.1121, nss: 1.7189, sauc: 0.6984, ig: 0.6877, aucj: 0.8504, aucb: 0.8205
    
    \hline
    \multicolumn{12}{c}{Backbone Networks} \\
    \hline
    ResNet34\_D64\_S\_AS &0.877&0.767&0.364&0.959&1.850&\textcolor{blue}{0.724}&0.851&0.863&\textcolor{blue}{0.833}&23.6&\textcolor{green}{131.0}\\% with recons 62.18, without recons 130.97. cc: 0.8767, sim: 0.7672, kld: 0.3642, emd: 0.9593, nss: 1.8504, sauc: 0.7241, ig: 0.8512, aucj: 0.8628, aucb: 0.8332
    ResNet18\_D64\_S\_AS &0.864&0.756&0.392&1.014&1.821&0.720&0.819&0.860&0.829&\textcolor{blue}{13.5}&\textcolor{red}{174.8}\\  %with recons 72.43, without recons 174.79. cc: 0.8643, sim: 0.7558, kld: 0.3920, emd: 1.0136, nss: 1.8207, sauc: 0.7203, ig: 0.8194, aucj: 0.8603, aucb: 0.8292
    %Res18\_64\_POET &0.797&0.702&-&-&1.649&-&-&0.847&0.808&&&\\
    %Res18\_64\_CAT2000 &0.863&0.754&-&-&1.821&-&-&0.862&0.829&&&\\
    %ConvNet\_64 &0.749&0.661&-&-&1.522&-&-&0.838&0.808&3.13&1.82&76.44\\
    MobileNetv3\_D64\_S\_AS &0.869&0.754&0.347&1.025&1.832&0.716&0.812&0.860&0.826&\textcolor{green}{4.0}&83.8\\ % with recons 48.89, without recons 83.75. cc: 0.8689, sim: 0.7540, kld: 0.3469, emd: 1.0249, nss: 1.8315, sauc: 0.7158, ig: 0.8119, aucj: 0.8603, aucb: 0.8256
    ShuffleNetv2\_D64\_S\_AS &0.867&0.751&\textcolor{green}{0.297}&1.038&1.820&0.718&0.811&0.861&0.831&\textcolor{red}{3.9}&99.9\\ % with recons 53.37, without recons 99.90. cc: 0.8674, sim: 0.7513, kld: 0.2966, emd: 1.0376, nss: 1.8201, sauc: 0.7179, ig: 0.8105, aucj: 0.8611, aucb: 0.8306
    
    \hline
    \end{tabular}
    \vspace{2mm}
    \caption{\small Ablation study on the SALICON validation set. The notation of method names follows \enquote{Backbone\_Scale\_Gaussian Covariance\_Grid Map}, down-sampled predictions by 32, 64 and 128$\times$ are   \enquote{\_D32}, \enquote{\_D64} and \enquote{\_D128}.
    We evaluate the use of isotropic Gaussians, on-diagonal variances only, and full covariances: (i) \enquote{\_S} ($\sigma_u\!=\!\sigma_v, \sigma_{u,v}\!=\!0$), (ii) \enquote{\_D} ($\sigma_{u,v}\!=\!0$) and (iii) \enquote{\_F}($\sigma_{u,v}, \sigma_u, \sigma_v$). We also evaluate \enquote{Anchor Settings}, including square grid (AS), horizontal-only grid (AH), vertical-only grid (AV), and none (AN).
    }
    \vspace{-3mm}
    \label{tab:ablation}
    \end{center}
\end{table*}

\subsection{Performance evaluation}
\label{performance_eva}

\noindent\textbf{Quantitative Evaluation.} We compare the proposed network with competing methods on three publicly available datasets and show the performance in Table~\ref{tab:salicon-test-performance}, Table~\ref{tab:mit1003-performance} and Table~\ref{tab:toronto_performance} respectively. The code and  models of several competing methods are available online, including Itti$^{\textcolor{red}{1}}$ and GBVS\footnote{http://www.klab.caltech.edu/~harel/share/gbvs.php}, OpenSALICON\footnote{https://github.com/CLT29/OpenSALICON}, ML-Net\footnote{https://github.com/marcellacornia/mlnet}, SalGAN\footnote{https://github.com/imatge-upc/salgan}, DeepGazeII\footnote{https://deepgaze.bethgelab.org},
SAM\footnote{https://github.com/marcellacornia/sam} and
MD-SEM\footnote{http://multiduration-saliency.csail.mit.edu}. We list nine evaluation metrics in these three tables. Our model SalGMM (ResNet50\_D64\_S\_AS in Table.~\ref{tab:salicon-test-performance}) uses
ResNet50 as the backbone.
\enquote{\textcolor{red}{red}} indicates the best performance, \enquote{\textcolor{green}{green}} is the second best and \enquote{\textcolor{blue}{blue}} is the third best. Comparing our performance with competing methods in all these three benchmark testing datasets, we observe that our model (\enquote{SalGMM}) ranks in the top three on most metrics.

\begin{figure*}
\centering
\includegraphics[width=0.92\linewidth]{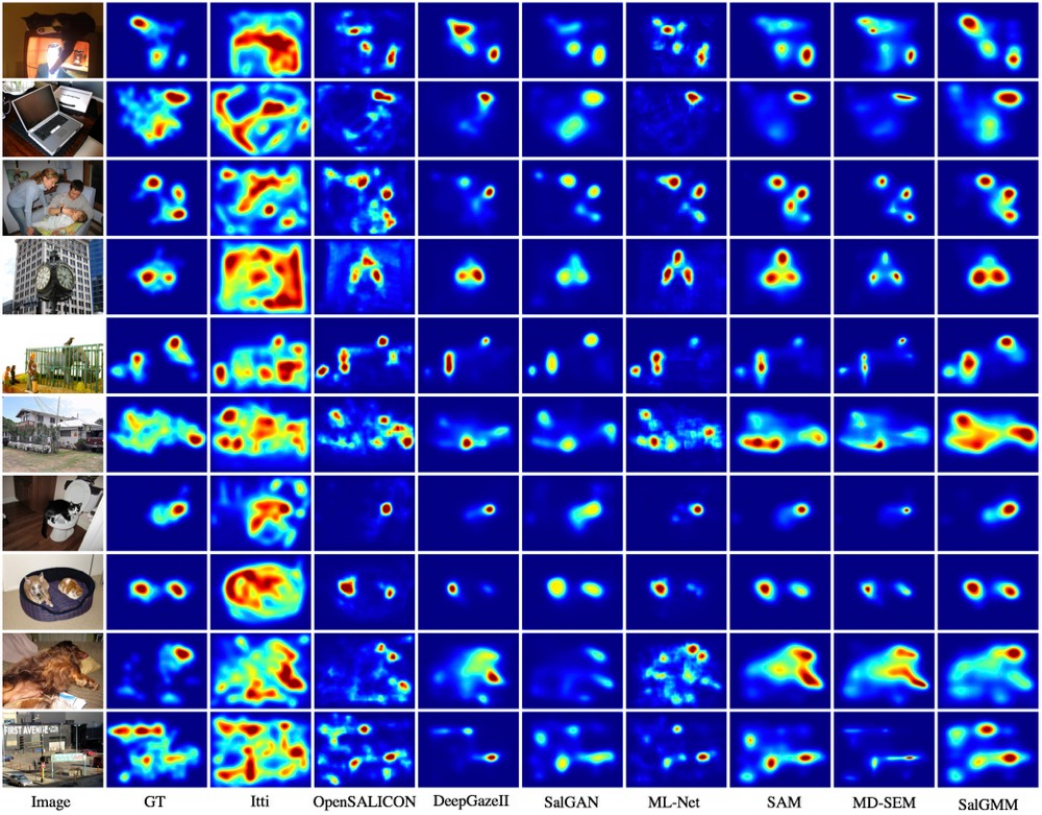}
\caption{Comparison of predictions for our method versus competing methods on the SALICON validation set.}
 \vspace{-3mm}
\label{fig:performance_compa_salicon_val}
\end{figure*}

\vspace{0.05cm}
\noindent\textbf{Qualitative Evaluation.}
Fig.~\ref{fig:performance_compa_salicon_val} shows several samples from the SALICON validation set on multiple competing methods, including Itti \cite{itti1998model}, OpenSALICON \cite{christopherleethomas2016}, DeepGazeII \cite{kummerer2017understanding}, SalGAN \cite{pan2017salgan}, ML-Net \cite{cornia2016deep}, SAM \cite{cornia2018predicting}, MD-SEM \cite{fosco2020much}. Itti is a hand-crafted method, while the others are deep models. Fig.~\ref{fig:performance_compa_salicon_val} shows that our model predicts a similar map compared to SAM and MD-SEM. However, our model performs better at locating the peak response of object regions. 

\vspace{0.05cm}
\noindent \textbf{Runtime and Model Size Comparison.}
We implement our GMM representation based eye fixation prediction model with multiple backbone networks as shown in Table.~\ref{tab:ablation}, including
ResNet (ResNet50, ResNet34, ResNet18) \cite{ResNet}, MobileNet-v3 \cite{howard2019searching} and ShuffleNet-v2 \cite{ma2018shufflenet}. The reported performance \enquote{SalGMM} in Table \ref{tab:salicon-test-performance}, Table \ref{tab:mit1003-performance} and Table \ref{tab:toronto_performance} corresponds to \enquote{ResNet50\_D64\_S\_AS} in Table \ref{tab:ablation}.
We observe the competing performance of our models compared with state-of-the-art models. Meanwhile, the smaller model size and higher frame/second rate further indicate both effectiveness and efficiency of our proposed solution.

We further show the inference time of six benchmark eye fixation models (ML-Net \cite{cornia2016deep}, DeepGazeII \cite{deepgaze2}, SAM \cite{cornia2018predicting}, OpenSalicon \cite{christopherleethomas2016}, SalGAN \cite{pan2017salgan} and MD-SEM \cite{fosco2020much}) and our models in Fig.~\ref{fig:speed_size}. The figure shows that all the competing methods have much longer inference time than ours. Meanwhile, the size of the circle indicates the size of the model. Our ShuffleNet and MobileNet based models have very small model sizes and comparable performance to the state-of-the-art methods.
Although ML-Net and DeepGazeII are smaller in model size than our ResNet50 and ResNet34 based models, the big gap in inference time further demonstrates the efficiency of our solution.

\subsection{Ablation Study}
We carried out the following ablation study to thoroughly analyze the proposed framework.

\vspace{0.05cm}
\noindent \textbf{Down Sampling Scales.} As mentioned in Section 3.2, we test 3 different scales of the down-sampling step.
The sub-block \enquote{Down sampling scales} in Table~\ref{tab:ablation} shows the performance on the SALICON validation set with the three different scales of down-sampling. The better performance of \enquote{ResNet50\_D32\_S\_AS} and \enquote{ResNet50\_D64\_S\_AS} compared with \enquote{ResNet50\_D128\_S\_AS} indicates that less down-sampling
works better in our scenario.

\vspace{0.05cm}
\noindent \textbf{Gaussian Covariance.} We evaluate
three modes of Gaussian covariance: \enquote{Spherical}, \enquote{Diag} and \enquote{Full}. \enquote{Spherical} represents an isotropic Gaussian. The level sets of \enquote{Spherical} Gaussian distribution are circles. For the \enquote{Diag} Gaussian distribution, the two elements $u$ and $v$ are independent. For the \enquote{Full} Gaussian distribution, $u$ and $v$ are correlated. The level sets of \enquote{Full} and \enquote{Diag} Gaussian distributions are ellipses. Table.~\ref{tab:ablation} shows the performance of the three types of Gaussian modes in the sub-block \enquote{Gaussian covariance}. The overall performance of \enquote{Diag} Gaussian distribution works better than the \enquote{Full} Gaussian distribution. 

\vspace{0.05cm}
\noindent \textbf{Anchor Settings.} In Fig.~\ref{fig:grid}, we show three anchor settings. The neural network computes the coordinate offsets from the anchor point  (marked as red points) to the ground-truth point rather than the location itself. The square grid is utilized in object detection networks \cite{ren2015faster}. 
We use both the horizontal grid and vertical grid map which is a popular practice in bag-of-words \cite{yang2010efficient, koniusz2016higher}. We observe that the square grid-based model (\enquote{ResNet50\_D64\_S\_AS}) outperforms the other ``horizontal only'', ``vertical only'' and ``no grid'' map, which indicates the square grid map works best in our task.

\vspace{0.05cm}
\noindent \textbf{Backbone Networks.}
As discussed before, we evaluate our solution for multiple backbone networks,
including ResNet (ResNet50, ResNet34, ResNet18) \cite{ResNet}, MobileNet-v3 \cite{howard2019searching} and ShuffleNet-v2 \cite{ma2018shufflenet}, and show the performance in Table \ref{tab:ablation}. The smaller size of MobileNet-v3 and ShuffleNet-v2 based models
allows our solution to be effectively applied on mobile devices.

\section{Conclusions}
In this paper, we formulate the 
eye fixation prediction as learning of parameters of Gaussian Mixture Model. Our model generates 216 output variables ($H \times W \times K$) whereas deep models that estimate a dense saliency map output 123,904 output variables with
an input size of $352\times352$. For the model size, learning a GMM instead of a dense saliency map requires a relatively small number of network parameters, \eg, {the parameters number of the model with ShuffleNet-v2 and MobileNet-v3 as backbones are 4M and 3.9M respectively.}
The new formulation of eye fixation also greatly decreases  
the inference time. We can reach a processing speed of 174.8 FPS, which indicates that our method works in real-time.
Further, we show that our approach leads to a smaller network with competitive
performance, which makes it 
suitable for embedded devices.
To the best of our knowledge, this paper is the first to formulate an eye fixation prediction task as GMM parameter estimation. We evaluated the proposed method on three public datasets: SALICON, MIT1003, and TORONTO. The performance comparison, inference time comparison, and network size comparison demonstrate that our method is fast and effective.

{\small
\bibliographystyle{ieee_fullname}
\bibliography{Eye_Fixation}
}

\end{document}